\documentclass[10pt,twocolumn,letterpaper]{article}

\usepackage{iccv}
\usepackage{times}
\usepackage{epsfig}
\usepackage{graphicx}
\usepackage{amsmath}
\usepackage{amssymb}
\usepackage[accsupp]{axessibility}

\usepackage{pdfpages}
\usepackage{multido}

\usepackage{booktabs}
\usepackage{bbding}
\usepackage{animate}
\usepackage{grffile}
\usepackage{siunitx} %
\usepackage{enumitem}
\usepackage[table, usenames, dvipsnames]{xcolor}
\usepackage{amsmath}
\usepackage{amssymb}
\usepackage{booktabs}
\usepackage{makecell}
\usepackage{multirow}
\usepackage{tabularx}
\usepackage{bbm}
\usepackage{caption} 
\usepackage{animate}
\usepackage{pifont}     %
\usepackage{xcolor}

\usepackage[breaklinks=true,letterpaper=true,colorlinks,bookmarks=false]{hyperref}

\newcommand{\xmark}{\ding{55}}
\newcommand{\cmark}{\ding{51}}%

\usepackage[capitalize]{cleveref}
\crefname{section}{Sec.}{Secs.}
\Crefname{section}{Section}{Sections}
\Crefname{table}{Table}{Tables}
\crefname{table}{Tab.}{Tabs.}

\DeclareMathOperator{\diag}{diag}
\newcommand{\norm}[1]{\left\lVert#1\right\rVert}

\iccvfinalcopy % *** Uncomment this line for the final submission

\newcommand{\notsosmall}{\fontsize{10.5pt}{12pt}\selectfont}

\def\iccvPaperID{4110} % *** Enter the ICCV Paper ID here

\definecolor{somegray}{rgb}{0.5, 0.5, 0.5}
\newcommand{\darkgrayed}[1]{\textcolor{somegray}{#1}}
\makeatletter
\newcommand*\titleheader[1]{\gdef\@titleheader{#1}}
\AtBeginDocument{%
  \let\st@red@title\@title
  \def\@title{%
    \vskip-3em
    \bgroup\normalfont\large\centering\@titleheader\par\egroup
    \vskip1.5em\st@red@title}
}
\makeatother

\titleheader{\darkgrayed{This paper has been accepted for publication at the \\
IEEE/CVF International Conference on Computer Vision (ICCV), Paris, 2023.
\copyright IEEE}}

\title{GO-SLAM: Global Optimization for Consistent 3D Instant Reconstruction}

\begin{document}

\author{Youmin Zhang \hspace{1.2cm} Fabio Tosi \hspace{1.2cm} Stefano Mattoccia \hspace{1.2cm}  Matteo Poggi\\
\notsosmall Department of Computer Science and Engineering (DISI) \\
\notsosmall University of Bologna, Italy \\
{\tt\small\{youmin.zhang2, fabio.tosi5, stefano.mattoccia, m.poggi\}@unibo.it} \\
\normalsize\url{https://youmi-zym.github.io/projects/GO-SLAM/}
}

\twocolumn[{%
\renewcommand\twocolumn[1][]{#1}%
\maketitle
\vspace{-1.0em}
   \includegraphics[width=\textwidth]{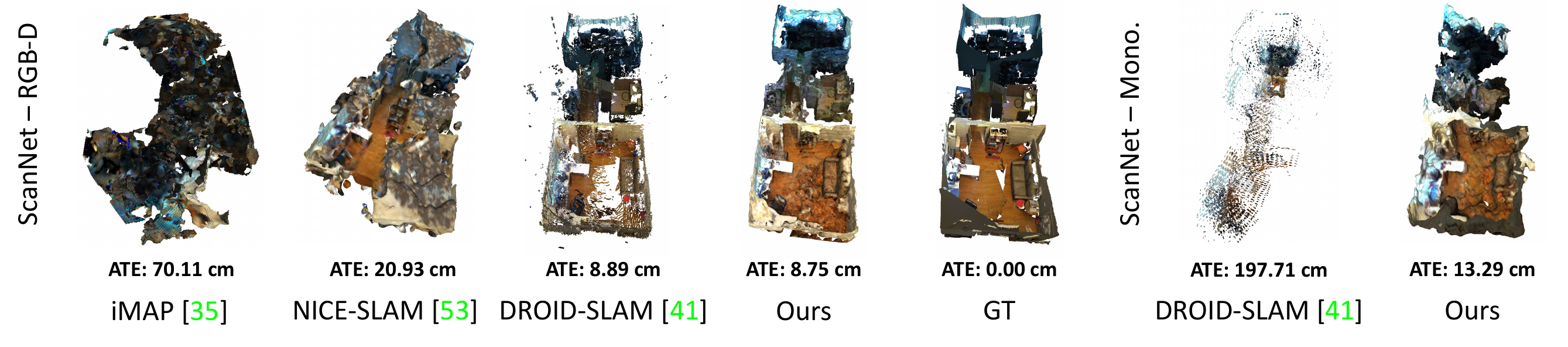}  
   \vspace{-0.6cm}
   \captionof{figure}{\small\textbf{3D Reconstruction and trajectory error on \textit{scene0054\_00} (ScanNet~\cite{scannet}).} 
   From left to right: RGB-D methods (iMAP~\cite{imap}, NICE-SLAM~\cite{niceslam}, DROID-SLAM~\cite{droidslam} and ours), ground truth scan, and monocular methods (DROID-SLAM~\cite{droidslam} and ours).}
  \label{fig:teaser}
  \vspace{0.4cm}
}]

\begin{abstract}

Neural implicit representations have recently demonstrated compelling results on dense Simultaneous Localization And Mapping (SLAM) but suffer from the accumulation of errors in camera tracking and distortion in the reconstruction. Purposely, we present GO-SLAM, a deep-learning-based dense visual SLAM framework globally optimizing poses and 3D reconstruction in real-time. Robust pose estimation is at its core, supported by efficient loop closing and online full bundle adjustment, which optimize per frame by utilizing the learned global geometry of the complete history of input frames. Simultaneously, we update the implicit and continuous surface representation on-the-fly to ensure global consistency of 3D reconstruction. Results on various synthetic and real-world datasets demonstrate that GO-SLAM outperforms state-of-the-art approaches at tracking robustness and reconstruction accuracy. Furthermore, GO-SLAM is versatile and can run with monocular, stereo, and RGB-D input.

\end{abstract}
\section{Introduction}
The demand for high-fidelity 3D object and scene reconstructions grows continuously in various fields, including robotics and augmented/virtual reality applications. Thus, faithfully representing objects and scenes in three-dimensional space is paramount to modelling them as continuous surfaces rather than discrete points. However, despite the significant advancements in 3D reconstruction techniques, obtaining high-quality representations in real-time without compromising accuracy and spatial resolution remains challenging. This fact is further demanding in online reconstruction scenarios, where handling camera motions and achieving real-time performance are critical.

Dense visual Simultaneous Localization and Mapping (SLAM) systems~\cite{kinectfusion,elasticfusion,badslam,bundlefusion,hrbf} have been introduced recently, enabling real-time, dense indoor scene reconstructions using RGB-D sensors. 
In particular, BundleFusion~\cite{bundlefusion} is the first volumetric approach that focuses on globally consistent 3D reconstruction on large-scale scenes at a real-time rate. However, consumer depth sensors have a limited working range~\cite{badslam,scannet} and could yield extremely noisy~\cite{nyuv2} measurements. These issues make the representation mapped by RGB-D SLAM suffer from blurring or over-smoothed geometric details, degrading the accuracy of pose estimation and reconstruction. 
In parallel, scene reconstruction from monocular imagery is emerging as a more convenient solution compared to RGB-D or LiDAR sensors. Camera sensors are lightweight, inexpensive, and represent the most straightforward configuration. 
Several deep-learning approaches~\cite{cnnslam,codeslam,deepfactors,deeptam,deepv2d,badslam,droidslam} have advanced monocular 3D reconstruction. However, their surface representations -- point cloud, surfel-based and volumetric representations -- lack flexibility at shape extraction and thus inhibit high-fidelity reconstruction.

More recently, the advent of Neural Radiance Fields (NeRFs) also impacted dense visual SLAM, offering photometrically accurate 3D representations of the world. The implicit representation yielded by continuous radiance fields allows for high-quality rendering of both visible and occluded regions, enabling extraction of the underlying shapes at arbitrary resolution.
Recent \cite{imap,di-fusion,niceslam} and concurrent \cite{li2023dense,nerfslam,orbeezslam,nicerslam} works demonstrate that NeRF-based visual SLAM can yield precise 3D reconstructions and camera pose estimation in small-scale scenes. However, due to the lack of global online optimization, such as loop closure (LC) and global bundle adjustment (BA), camera drift error accumulates as the number of processed frames grows, and the 3D reconstruction quickly collapses, as shown in \cref{fig:teaser}.

Purposely, this paper introduces GO-SLAM, a deep-learning-based SLAM system featuring on-the-fly, globally consistent 3D reconstruction, facilitated by our robust camera tracking and real-time implicit surface updates. As real-world 3D scenes may be very complex, drifting cannot be completely avoided by only locally tracking camera motion, especially in the monocular camera setting, due to the lack of explicit depth measurements. In addition to the local registration commonly performed by current SLAM systems, we present an efficient loop closing to correct trajectory in real-time, accompanied by an online full BA module to actively optimize the 3D geometry of the complete keyframes history. In contrast to previous works performing BA or LC with sparse visual features~\cite{orbslam2,orbslam3,bundlefusion,deepfactors}, our end-to-end global optimization procedure is naturally robust to challenging regions, thanks to richer features and geometry cues (\eg, pixel-wise flow) learned by neural networks. Furthermore, GO-SLAM implements instant mapping based on a neural implicit network with multi-resolution hash encoding~\cite{instantngp}. Its compact, multiscale representation enables updating the 3D reconstruction at high-frequency according to newly-optimized camera poses and depths from our global optimization system, thus ensuring global consistency in the dense map and capturing local details. 
Our contributions can be resumed as follows:

\begin{itemize}
    \item A novel deep-learning-based, real-time global pose optimization system that considers the complete history of input frames and continuously aligns all poses.

    \item An efficient alignment strategy that enables instantaneous loop closures and correction of global structure, being both memory and time efficient.

    \item  An instant 3D implicit reconstruction approach, enabling on-the-fly and continuous 3D model update with the latest global pose estimates. This strategy facilitates real-time 3D reconstructions.

    \item The first deep-learning architecture for joint robust pose estimation and dense 3D reconstruction suited for any setup: monocular, stereo, or RGB-D cameras. 
\end{itemize}

\section{Related Work}

Here, we review the literature relevant to our work.

\textbf{Online 3D Reconstruction and SLAM.} Real-time, dense, and globally consistent 3D reconstruction is crucial in the SLAM literature. As a core element for high-quality reconstruction, the underlying representation can be approximately categorized as depth point~\cite{orbslam,orbslam2,orbslam3,deeptam,svo,codeslam,cnnslam,deepfactors,droidslam}, height map~\cite{gallup20103d}, surfel~\cite{badslam,elasticfusion} and volumetric representation~\cite{kinectfusion,voxelhashing,bundlefusion}. Especially some of them~\cite{orbslam2,orbslam3,bundlefusion,badslam,deepfactors}, make an effort for globally consistent reconstruction by implementing global BA and LC systems. 
However, due to discrete and limited surface representations (\eg, point-, surfel- or voxel-based), these methods suffer from the accumulation of errors in camera tracking and distortion in the reconstruction. DROID-SLAM~\cite{droidslam} achieves impressive trajectory estimations by using neural networks to leverage richer context from images, yet it performs global bundle adjustment only offline, at the end of camera tracking. 
However, for some challenging cases, it gets hard to eliminate the drift error with offline refinement solely, as shown in \cref{fig:teaser}, pointing out the importance of online BA for on-the-fly drift correction. 

\begin{figure*}[t]
    \centering
    \includegraphics[width=1.96\columnwidth]{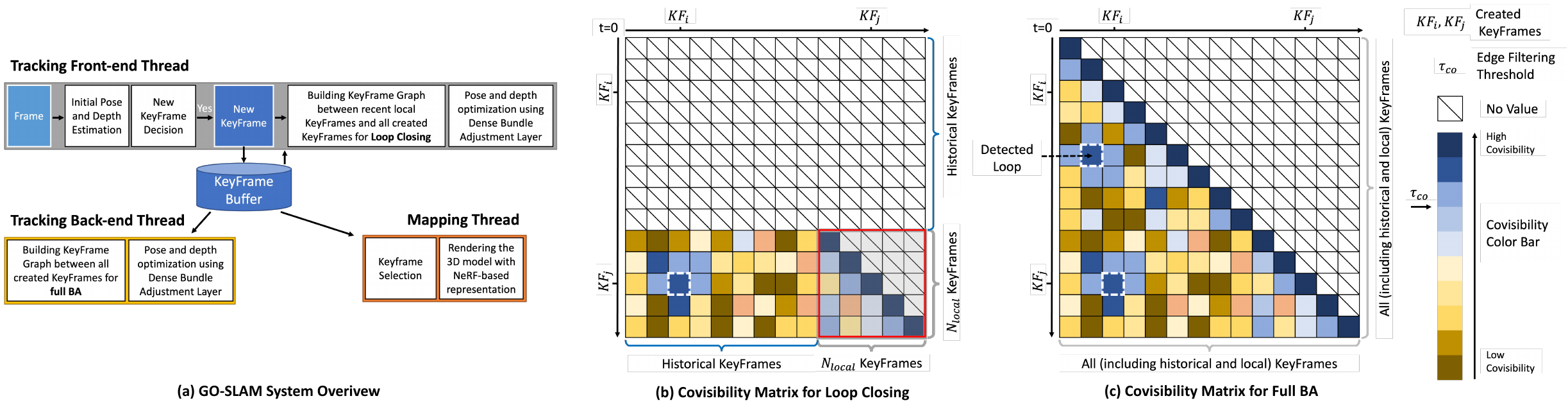}
    \vspace{-0.3cm}
    \caption{\textbf{Architecture Overview.} Our GO-SLAM framework consists of three parallel threads (a): front-end tracking (including keyframe initialization and loop closing), back-end tracking, and instant mapping. 
    The front-end tracking thread uses the video stream as input and iteratively updates the pose and depth of the current frame while determining whether it should be promoted as a new keyframe. Moreover, it also actively performs efficient loop closing (b). The back-end tracking thread focuses on generating globally consistent pose and depth predictions through full bundle adjustment (c). Simultaneously, instant mapping updates the 3D reconstruction on-the-fly according to the latest geometry changes.}
    \label{fig:framework}
\end{figure*}

\textbf{NeRF-based Visual SLAM.}
Neural implicit fields (NeRF) have recently emerged as one of the promising and widely applicable methods for 3D representation, opening up many new research opportunities including novel view synthesis~\cite{nerf,ibrnet,neus,mipnerf,mipnerf360,nerf++,dvgo,plenoxels,instantngp}, multi-view 3D reconstruction~\cite{mvsnerf,rcmvsnet,neural3dwild}, and large-scale scene reconstruction~\cite{gosurf,mononeuralfusion,bnvfusion,nerfusion,isdf,neuralsurface,Yu2022MonoSDF}. A common requirement of these methods is the posed images. \cite{barf,inerf} tries to relax this constraint starting from imperfect camera poses on small objects. More recently, using neural implicit representation in visual SLAM~\cite{imap,niceslam,nicerslam,nerfslam,orbeezslam,li2023dense} has achieved better scene completeness, especially for unobserved regions, and it has also allowed for continuous 3D modeling at arbitrary resolution. Two pioneer works, iMAP~\cite{imap} and NICE-SLAM~\cite{niceslam}, extend the neural implicit representation to RGB-D SLAM system, which learns camera pose tracking and room-scale mapping from scratch. Concurrent works \cite{li2023dense,nicerslam}, by removing the reliance on depth sensor input, achieve visual SLAM when only RGB sequences are available. Instead of na\"ive pose optimization with NeRF, Orbeez-SLAM~\cite{orbeezslam} resorts to visual odometry from ORB-SLAM2~\cite{orbslam2} for accurate pose estimation. However, the lack of many of the core capabilities of modern SLAM systems, such as LC and global BA, inhibits the ability of these methods to perform large-scale reconstructions. Concurrent to our work, NeRF-SLAM~\cite{nerfslam} integrates DROID-SLAM~\cite{droidslam} for camera tracking. However, it also inherits its limitations, \ie, the absence of online loop closing and full BA, which restricts its ability to perform globally consistent 3D reconstruction.

\section{Method}

Our GO-SLAM framework, depicted in \cref{fig:framework}, uses a keyframe-based SLAM paradigm to achieve real-time, globally-consistent 3D reconstruction. This is made possible by the online drift-corrected pose tracking and instant mapping capabilities of our system. By performing full bundle adjustment and loop closing, pose optimization can be carried out globally. Meanwhile, instant mapping adapts to continuous changes in optimized global poses and depths.

\subsection{Tracking with Global Optimization}
\begin{figure*}[!t]
    \centering
    \includegraphics[width=2.0\columnwidth]{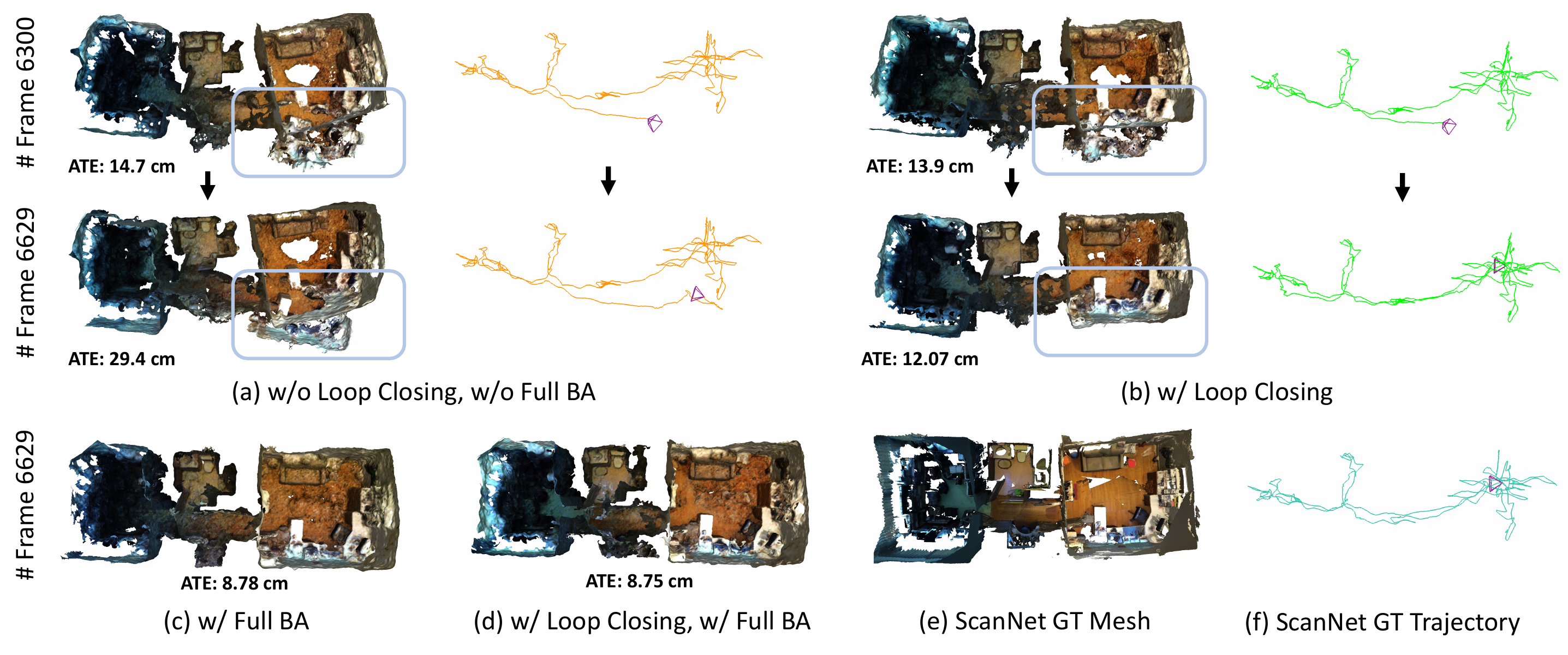}
    \caption{\textbf{Qualitatives examples of LC and full BA on \emph{scene0054\_00} (ScanNet~\cite{scannet}) with a total of 6629 frames.} In (a), a significant error accumulates when no global optimization is available. With loop closing (b), the system is able to eliminate the trajectory error using global geometry. Additionally, online full BA optimizes (c) the poses of all existing keyframes. The final model (d), which integrates both loop closing and full BA, achieves a more complete and accurate 3D model prediction.}
    \label{fig:ba}
\end{figure*}

Loop closing and global bundle adjustment are crucial for robust pose estimation and long-term map consistency. In this work, we extend the tracking component of DROID-SLAM~\cite{droidslam} by several key features. 
These enhancements effectively reduce the drift and pave the way for a globally optimal map. In the front-end tracking, we initialize a keyframe if sufficient motion is observed -- there, we also introduce LC -- while the back-end is equipped with full BA for online global refinement.
\cref{fig:ba} shows the effect of both LC and BA on tracking and 3D reconstruction qualitatively, correcting the large errors occurring in their absence.

\textbf{Front-End Tracking.} Our system takes as input a live video stream, which can be either monocular, stereo, or RGB-D, and applies a recurrent update operator based on RAFT ~\cite{raft} to compute the optical flow of each new frame compared to the last keyframe. If the average flow is larger than a predefined threshold $\tau_{flow}$, a new keyframe is created out of the current frame and added to the maintained keyframe buffer for further refinement. 

We use the set of keyframes $\{ \mathbf{KF}_{k} \}_{k=1}^{N_{KF}}$ created so far to build a keyframe-graph $(\mathcal{V}, \mathcal{E})$ for performing LC. This process involves two steps: (1) select high co-visibility connections between the most recent $N_{local}$ keyframes, and (2) detect loop closures between local keyframes and historical keyframes outside the local window. Accordingly, we compute a co-visibility matrix of size $N_{local} \times N_{KF}$ between $N_{local}$ local keyframes and all the $N_{KF}$ created keyframes to find valuable edge connections, as shown in \cref{fig:framework} (b). In practice, the co-visibility is represented by the mean rigid flow between keyframe pairs using efficient back-projection, and those with low co-visibility, \ie, mean flow larger than $\tau_{co}$, are filtered out. Among local keyframes -- the red-borders sub-matrix in \cref{fig:framework} (b) -- we build edges for keyframe pairs temporally adjacent or with high co-visibility. To avoid redundancy, once an edge connection (\eg, KF$_{i}\leftrightarrow$KF$_{j}$) is added to the keyframe-graph, we suppress all possible neighboring edges between $\{ \mathbf{KF}_{k} \}_{k=i-r_{local}}^{i+r_{local}}$ and $\{ \mathbf{KF}_{k} \}_{k=j-r_{local}}^{j+r_{local}}$, where $r_{local}$ is a hyper-parameter denoting a temporal radius. 
The loop detection step is quite similar. We sample edges from the unexplored part of the co-visibility matrix in descending order of co-visibility and suppress neighboring edges with radius $r_{loop}$. More strictly, to accept a loop candidate, we detect consecutively three loop candidates and validate them if their mean flow is lower than $\tau_{co}$. The value of $r_{loop}$ is determined empirically based on the observation that keyframes within a local window observe almost the same scene. We set $r_{loop}$ to $\frac{N_{local}}{2}$, which allows for only one loop closure between the recent local region and one revisited region. In addition to the edge connections within local keyframes, several extra loop edges can be added to the keyframe graph depending on how many times the current local region is revisited. In general, the number of edges in the graph is linear to $N_{local}$ with an upper-bound $N_{local}\times N_{local}$ $+$ few loop closures. Through neighborhood suppression and co-visibility filtering, we further limit the number of edges in the keyframe-graph to $s_{edge}\cdot N_{local}$, so that the efficiency of optimization of the entire keyframe graph, \ie, the whole front-end tracking, can be further ensured.

Afterward, we use the differentiable Dense Bundle Adjustment (DBA) layer proposed in ~\cite{droidslam} to solve a non-linear least squares optimization problem over the cost function in order to correct the camera pose $\mathbf{G} \in SE(3)$ and inverse depth $\mathbf{d} \in \mathbb{R}_{+}^{H \times W}$ of each keyframe in the keyframe-graph:
\begin{equation}
    \mathbf E(\mathbf{G}, \mathbf{d}) = \sum_{(i,j) \in \mathcal{E}} \norm{\mathbf{p}_{ij}^* - \Pi_c(\mathbf{G}_{ij}  \circ \Pi_c^{-1}(\mathbf{p}_i, \mathbf{d}_i)) }_{\Sigma_{ij}}^2,
\label{eq:objective}
\end{equation}
where $(i, j) \in \mathcal{E} $ denotes any edge in keyframe-graph, $\Pi_c$ and $\Pi_c^{-1}$ are the projection and back-projection functions, $\mathbf{p}_{i}$ is the back-projected pixel position from keyframe $\mathbf{KF}_i$, $\mathbf{G}_{ij}$ is the pose transformation from $\mathbf{KF}_i$ to $\mathbf{KF}_j$, $\mathbf{p}_{ij}^{*}$ and $\mathbf{w}_{ij}$ are estimated flow and associated confidence map, $\Sigma_{ij} = \diag \mathbf{w}_{ij}$ and $\norm{\cdot}_{\Sigma}$ is the Mahalanobis distance which weights the error terms based on the confidence weights $\mathbf{w}_{ij}$. The cost function allows the update of camera poses and dense per-pixel depth to maximize their compatibility with flow $\mathbf{p}_{ij}^{*}$ predicted by the recurrent update operator. For the sake of efficiency, we only compute the Jacobians with respect to the depths and poses of the local keyframes. After computing residuals and Jacobians at each iteration, a damped Gauss-Newton algorithm is applied to find the optimal poses and depths of all local keyframes.

\begin{figure}[!t]
    \centering
    \includegraphics[width=1.0\columnwidth]{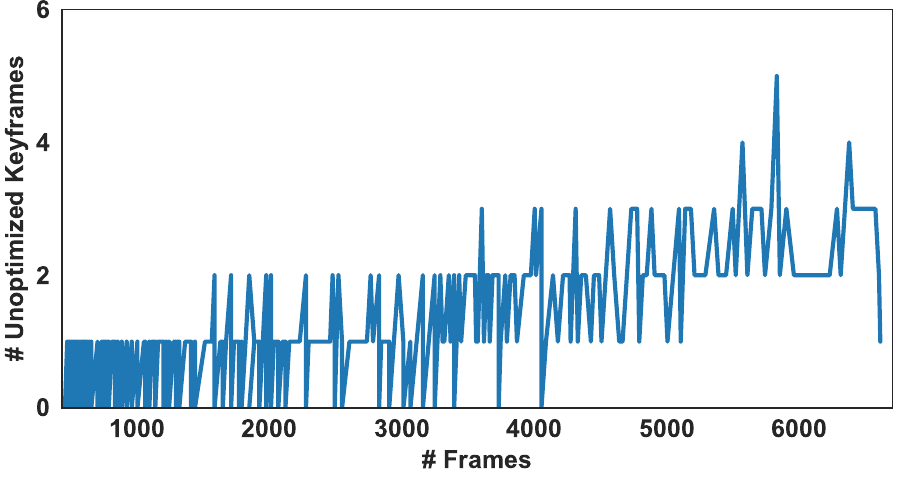}
    \vspace{-0.3cm}
    \caption{\textbf{Number of unoptimized keyframes at each timestamp with full BA.} Experiment on \emph{scene0054\_00} (ScanNet~\cite{scannet}) with 6629 frames (357 total keyframes). }
    \label{fig:full_ba}
\end{figure}

\textbf{Back-End Tracking.}
Simultaneously optimizing overall historical keyframes can be
computationally expensive, as observed in~\cite{orbslam2,orbslam3,droidslam}. To address this issue, we follow a similar approach as previous SLAM algorithms~\cite{orbslam2,orbslam3} by running the full BA online in a separate thread, allowing the system to continue tracking new frames and loop closing. Similarly to the proposed front-end tracking, we start a new keyframe-graph and insert keyframe pairs with high co-visibility, as well as temporal adjacent keyframes, as shown in \cref{fig:framework} (c). When a new edge is built, we suppress the redundant neighboring edges with radius $r_{global}$. As the trajectory error of the latest keyframes has been corrected with global geometry featured by loop closing, it eases the real-time requirement for the full BA. Our proposed full BA is efficient up to tens of thousands of input frames, as shown in \cref{fig:full_ba}.

\subsection{Instant Mapping}
The proposed instant mapping aims at updating the global 3D reconstruction in real-time by incorporating newly-optimized geometry from tracking. However, this goal presents two challenges for the mapping thread: i) ensuring that the updated reconstruction remains globally consistent, and ii) enabling fast rendering of the reconstructed scene.
Updating all existing keyframes at once is the simplest approach to ensure global consistency, but it can quickly become impractical as the number of keyframes increases. Therefore, it is crucial to prune the keyframe candidates for updating selectively. Additionally, high-speed rendering of the scene is necessary to meet real-time requirements. To achieve these goals, we introduce our novel keyframe selection strategy and discuss our use of spatial hashing~\cite{instantngp} and rendering networks in the remainder.

\textbf{Keyframe Selection.} At the beginning of each update of the 3D reconstruction, the instant mapping thread first takes a snapshot of all existing keyframe poses and depths tracked, to ensure that the geometry remains consistent during the mapping period. For keyframe selection, we prioritize those with the most relevant optimization updates, following the principle established by previous works~\cite{bundlefusion}. Firstly, we ensure that the latest two keyframes and those not optimized by mapping are always included. In addition, following~\cite{bundlefusion}, we sort all keyframes in descending order of pose difference between the current and last updated state and select the top 10 keyframes from the sorted list when accessing. Furthermore, to prevent the mapping from forgetting previous 3D geometry, we also select 10 keyframes using a stratified sampling~\cite{nerf} from all the available keyframes. 

\textbf{Rendering.} Drawing inspiration from the recent advancements in implicit neural network techniques~\cite{nerf} -- Instant-NGP~\cite{instantngp} in particular -- we can construct 3D models from scratch with remarkable speed and accuracy. Specifically, given depth $\mathbf{D}$ (converted from inverse depth $\mathbf{d}$), pose $\mathbf{G}$, and image $\mathbf{I}$ for each selected keyframe, we randomly select $M$ pixels for training. We then generate $N_{ray} = N_{strat} + N_{imp}$ total sampling points along the emitted ray crossing each pixel, where $N_{strat}$ points are sampled using stratified sampling and $N_{imp}$ points are selected near the depth value following \cite{niceslam}. For each 3D sampling point $\mathbf{x}$, we map it to multi-resolution hash encodings~\cite{instantngp} $h_{\Theta_{hash}}(\mathbf{x})$ with trainable encoding parameters $\Theta_{hash}$ at each entry of the hash table. As the hash encodings $h_{\Theta_{hash}}(\mathbf{x})$ have explicitly stored the geometric and intensity information for each spatial position, we can predict a signed distance function (SDF) $\Phi(\mathbf{x})$ and color $\Omega(\mathbf{x})$ using shallow networks. 

More specifically, our SDF network $f_{\Theta_{sdf}}$, which consists of a single multi-layer perceptron (MLP) with learnable parameters $\Theta_{sdf}$, takes the point position $\mathbf{x}$ and corresponding hash encodings $h_{\Theta_{hash}}(\mathbf{x})$ as input and predicts the SDF as:
\begin{align}
    \Phi(\mathbf{x}), \; \mathbf{g} = f_{\Theta_{sdf}}(\mathbf{x}, h_{\Theta_{hash}}(\mathbf{x})),
\label{eq:sdf}
\end{align}
with $\mathbf{g}$ being the learned geometry feature vector. 
The color network $f_{\Theta_{color}}$ processes $\mathbf{g}$, $\mathbf{x}$ and the gradient of SDF $\mathbf{n}$ with respect to $\mathbf{x}$ to estimate color $\Omega(\mathbf{x})$ as:
\begin{align}
    \Omega(\mathbf{x}) = f_{\Theta_{color}}(\mathbf{x}, \mathbf{n}, \mathbf{g}).
\label{eq:color}
\end{align}
where $\Theta_{color}$ is the set of learnable parameters of the color network, i.e., a two-layers MLP.

The depth and color of each pixel/ray are calculated by unbiased volume rendering following NeuS~\cite{neus}. Specifically, for point $\mathbf{x}_{i},\; i\in \{1, \cdots, N_{ray}\}$ along a ray, given the camera center $\mathbf{o}$, view direction $\mathbf{v}$, and sampled depth $D_{i}^{ray}$, it can be formulated as $\mathbf{x}_{i} = \mathbf{o} + D_{i}^{ray}\mathbf{v}$. At the same time, the unbiased ray termination probability at this point is modeled as $w_{i}=\alpha_{i}\Pi_{j=1}^{i-1}(1-\alpha_{j})$, where the opacity value $\alpha_{i}$ is computed as:
\begin{equation}
    \alpha_{i} = \max\left( \frac{\sigma(\Phi(\mathbf{x_{i}}))-\sigma(\Phi(\mathbf{x_{i+1}}))}{\sigma(\Phi(\mathbf{x_{i}}))}, 0 \right).
\label{eq:opacity}
\end{equation}
where $\sigma$ is the modulated Sigmoid function~\cite{neus}. With the weight $w$, the predictions of pixel-wise color $\hat{\mathbf{c}}$ and depth $\hat{\mathbf{D}}$ are accumulated along the ray:
\begin{align}
    \hat{\mathbf{c}} = \sum_{i=1}^{N_{ray}}w_{i}\Omega(\mathbf{x}_{i}), \quad \hat{\mathbf{D}} = \sum_{i=1}^{N_{ray}}w_{i}D_{i}^{ray}.
\label{eq:predition}
\end{align}

\begin{table*}[t]
\centering
\begin{tabular}{cc}
\resizebox{0.64\linewidth}{!}{%
\begin{tabular}{l|ccccccccc | c}
\toprule
& 360 & desk & desk2 & floor & plant & room & rpy & teddy & xyz & avg \\
\toprule
ORB-SLAM2~\cite{orbslam2} & \xmark & 0.071 & \xmark & 0.023 & \xmark & \xmark & \xmark & \xmark & 0.010 & - \\
ORB-SLAM3~\cite{orbslam3} & \xmark & 0.017 & 0.210 & \xmark & 0.034 & \xmark & \xmark & \xmark & \textbf{0.009} & - \\
\midrule
DeepV2D~\cite{deepv2d} & 0.243 & 0.166 & 0.379 & 1.653 & 0.203 & 0.246 & 0.105 & 0.316 & 0.064 & 0.375 \\
% DeepV2D (TartanAir) & 0.182 & 0.652 & 0.633 & 0.579 & 0.582 & 0.776 & 0.053 & 0.602 & 0.150 & 0.468 \\
DeepFactors~\cite{deepfactors} & 0.159 & 0.170 & 0.253 & 0.169 & 0.305 & 0.364 & 0.043 & 0.601 & 0.035 & 0.233\\
DROID-SLAM~\cite{droidslam} & 0.111 & 0.018 & 0.042 & \textbf{0.021} & 
\textbf{0.016} & 0.049 & 0.026 & \textbf{0.048} & 0.012 & 0.038 \\
\midrule
Ours & \textbf{0.089} & \textbf{0.016} & \textbf{0.028} & 0.025 & 0.026 & 0.052 & \textbf{0.019} & \textbf{0.048} & 0.010 & \textbf{0.035} \\

\bottomrule
\end{tabular}
}
& 
\resizebox{0.675\columnwidth}{!}{
\begin{tabular}{lccc}
\toprule
& fr1/desk &  fr2/xyz &  fr3/office \\
\midrule
{Kintinuous}\cite{kinectfusion} & 0.037 &  0.029  & 0.030 \\
{BAD-SLAM}\cite{badslam} & 0.017  & 0.011  & 0.017 \\
{ORB-SLAM2}\cite{orbslam2} & 0.016  & \bf 0.004  & \bf 0.010 \\
\midrule
{iMAP}~\cite{imap} & 0.049 & 0.020 & 0.058  \\
{NICE-SLAM~\cite{niceslam}} & 0.027 & 0.018 & 0.030 \\
\midrule
GO-SLAM (ours) & \bf 0.015 & 0.006 & 0.013 \\
\bottomrule
\end{tabular}
}
\end{tabular}
\caption{\textbf{ATE[m] on the TUM RGB-D~\cite{tum} benchmark.} The left table shows the results of \textbf{monocular} SLAM on sequences from the freiburg1 set. The right one reports the accuracy for \textbf{RGB-D} SLAM on sequences from freiburg1, freiburg2 and freiburg3 respectively. `\xmark’ denotes tracking failure, `-’ no available data. Results of monocular SLAM~\cite{orbslam2,orbslam3,deepv2d,deepfactors,droidslam} and RGB-D SLAM~\cite{kinectfusion,badslam,orbslam2,imap,niceslam} are taken from \cite{droidslam} and \cite{niceslam} respectively.}
\label{table:TUM}
\end{table*}

\textbf{Training Losses.} To optimize the rendering networks, taking keyframe image $\mathbf{I}$ and depth $\mathbf{D}$ as ground truth, the RGB and depth losses $\mathcal{L}_{c}$ and $\mathcal{L}_{dep}$ imposed on the selected $M$ pixels are $\mathcal{L}_{c} = \frac{1}{M} \sum_{m=1}^{M} |\mathbf{c}_{m} - \hat{\mathbf{c}}_{m}|,\; \mathbf{c}_{m} \in \mathbf{I} $ and 
\begin{align}
    \mathcal{L}_{dep} = \frac{1}{M} \sum_{m=1}^{M} \frac{|\mathbf{D}_{m} - \hat{\mathbf{D}}_{m}|}{\sqrt{\hat{\mathbf{D}}_{m}^{var}}}, 
\label{eq:depthloss}
\end{align}
respectively, with $\hat{\mathbf{D}}^{var}_{m} = \sum_{i=1}^{N_{ray}}w_{i}(\hat{\mathbf{D}}_{m} - D_{m,i}^{ray})^{2}$ being the predicted depth variance used for down-weighting uncertain regions in the reconstructed geometry~\cite{imap,niceslam}. Furthermore, we also introduce regularization to the predicted SDF, following~\cite{isdf,gosurf}. Specifically, to encourage the gradient of SDF to unit length, we adopt the Eikonal term~\cite{eikonal}:  
\begin{equation}
    \mathcal{L}_{eik}=\frac{1}{MN_{ray}}\sum_{m,i}(1-||\mathbf{n}_{m, i}||)^{2}
\end{equation}
 Besides, to supervise the SDF for accurate surface reconstructions, we approximate the ground truth SDF of sampling point $\mathbf{x}_{i}$ by computing its distance to the keyframe's depth $\mathbf{D}_{m}$, \ie, $\mathbf{b}(\mathbf{x}_{i}) = \mathbf{D}_{m} - D_{m,i}^{ray}$. For SDF learning, we have $|\Phi(\mathbf{x}_{i})| \leq |\mathbf{b}(\mathbf{x}_{i})|, \forall \mathbf{x}_{i}$. To satisfy this bound, for near-surface points ($|\mathbf{b}(\mathbf{x}_{i})| \leq \tau_{trunc}$, where $\tau_{trunc}$ is a hyper-parameter denoted the truncation threshold, set to 16cm), the SDF loss is defined as $\mathcal{L}_{near} = |\Phi(\mathbf{x}_{i}) - \mathbf{b}(\mathbf{x}_{i})|$, while for elsewhere, \ie, free space, we apply a relaxed loss:
\begin{align}
    \mathcal{L}_{free} = \max\left(e^{-\beta\Phi(\mathbf{x}_{i})} - 1, \Phi(\mathbf{x}_{i}) - \mathbf{b}(\mathbf{x}_{i}), 0 \right)
\end{align}
where $\beta$ is a hyper-parameter to apply a penalty when the predicted SDF $\Phi(\mathbf{x}_{i})$ is negative in free space. Therefore, our full SDF loss is defined as:
\begin{equation}
    \mathcal{L}_{sdf} = \frac{1}{MN_{ray}}\sum_{m,i} \left\{ \begin{array}{ll}
       \mathcal{L}_{near} & \textbf{if } |\mathbf{b}(\mathbf{x}_{i})| \leq \tau_{trunc} \\
        \mathcal{L}_{free} & \textbf{otherwise.} 
    \end{array} \right.
\end{equation}

Our instant mapping thread continuously optimizes the scene reconstruction over all sampled pixels of all selected keyframes. Specifically, for each set of selected keyframes, we run the mapping process for a fixed number $N_{iter}$ of iterations. Our total loss is defined as:
\begin{align}
    \mathcal{L} = \lambda_{c} \mathcal{L}_{c} + \lambda_{dep} \mathcal{L}_{dep} + \lambda_{eik} \mathcal{L}_{eik} + \lambda_{sdf} \mathcal{L}_{sdf},
\end{align}
with $\mathcal{L}_{c}$, $\mathcal{L}_{dep}$, $\mathcal{L}_{eik}$, and $\mathcal{L}_{sdf}$ being loss balance weights.
Given the pose and depth of each selected keyframe, our mapping thread directly uses them without refinement since our full BA and LC have exploited the global geometry to optimize both, being naturally robust in handling occluded regions and back faces.  
\section{Experimental Results}

This section investigates our experimental evaluation, including implementation details, datasets, and key findings.

\subsection{Implementation Details}
Our system runs on a PC with a 3.5GHz Intel Core i9-10920X CPU and an NVIDIA RTX 3090 GPU. For tracking, we utilize pre-trained weights from DROID-SLAM~\cite{droidslam}, whereas the rendering networks are trained from scratch. The experiments are performed with the following default settings unless otherwise specified: local window size $N_{local} = 25$ for RGB-D and stereo input, $N_{local} = 50$ for monocular mode, neighboring radius $r_{local}=1$, $r_{global}=5$, co-visibility threshold $\tau_{co}=25.0$, the factor used to constrain the maximum allowed edge $s_{edge}=8$, sampling points along a ray $N_{strat}=24$, $N_{imp}=48$, pixel samples $M=200$, penalty parameter $\beta=5.0$, iterations $N_{iter}=2$, and the loss weights $\lambda_{c}=1.0$, $\lambda_{dep}=1.0$, $\lambda_{eik}=0.1$, $\lambda_{sdf}=1.0$. The mesh reconstruction of a scene is achieved by running marching cubes on the SDF values of the queried points. The supplementary material provides more details about SLAM configuration and also qualitative results on various datasets.

\subsection{Datasets}
We evaluate our GO-SLAM system on several datasets with different input modalities, including the TUM RGB-D, EuRoC, ETH3D-SLAM, ScanNet, and Replica. The TUM RGB-D benchmark~\cite{tum} is a small-scale indoor dataset with accurate ground truth obtained from an external camera motion capture system. 
The EuRoC dataset ~\cite{euroc} contains 11 indoor stereo sequences recorded from a micro aerial vehicle (MAV). The ETH3D-SLAM dataset~\cite{badslam} provides real-world RGB-D image sequences captured with synchronized global shutter cameras. The ScanNet dataset \cite{scannet} features richly annotated RGB-D scans of real-world environments, including challenging short and long trajectories. Finally, the Replica dataset \cite{replica} provides high-fidelity 3D models of photo-realistic indoor scenes, enabling us to assess the reconstruction performance of our approach. For TUM RGB-D, EuRoC, and ETH3D-SLAM, images are resized to $384\times 512$ resolution, while $240 \times 320$ and $320 \times 640$ for ScanNet and Replica datasets, respectively.

\begin{table*} [t]
\centering
\begin{tabular}{cc}
\resizebox{0.7\linewidth}{!}{%
\begin{tabular}{l| ccccc | ccc | ccc | c}
\toprule
& MH01 & MH02 & MH03 & MH04 & MH05 & V101 & V102 & V103 & V201 & V202 & V203 & Avg \\
\toprule
ORB-SLAM2~\cite{orbslam2} & 0.035 & 0.018 & 0.028 & 0.119 & 0.060 & \textbf{0.035} & 0.020 & 0.048 & 0.037 & 0.035 & - & - \\
SVO~\cite{svo} & 0.040 & 0.070 & 0.270 & 0.170 & 0.120 & 0.040 & 0.040 & 0.070 & 0.050 & 0.090 & 0.790 & 0.159 \\
ORB-SLAM3~\cite{orbslam3} & 0.029 & 0.019 & 0.024 & 0.085 & 0.052 & \textbf{0.035} & 0.025 & 0.061 & 0.041 & 0.028 & 0.521 & 0.084 \\
DROID-SLAM~\cite{droidslam} & \textbf{0.015} & \textbf{0.013} & 0.035 & 0.048 & \textbf{0.040} & 0.037 & \textbf{0.011} & \textbf{0.020} & 0.018 & 0.015 & \textbf{0.017} & \textbf{0.024} \\

\midrule
Ours & 0.016 & 0.014 & \textbf{0.023} & \textbf{0.045} & 0.045 & 0.037 & \textbf{0.011} & 0.023 & \textbf{0.016} & \textbf{0.010} & 0.022 & \textbf{0.024}  \\

\bottomrule
\end{tabular}
}
&
\resizebox{0.52\columnwidth}{!}{
\begin{tabular}{lccc}
\toprule
& V103 & V202 & V203 \\
\midrule
ORB-SLAM3~\cite{orbslam3} & 0.037 & 0.022 & \xmark \\ 
DSO~\cite{dso} & 0.903 & 0.132 & 1.152 \\
DROID-SLAM~\cite{droidslam} & 0.020 & 0.013 & \bf 0.014 \\
\midrule
Li \emph{et al.}~\cite{li2023dense} & \xmark & 0.178 & \xmark \\
\midrule
GO-SLAM (ours) & \bf 0.018  & \bf 0.011 & 0.017 \\
\bottomrule
\end{tabular}
}
\end{tabular}
\caption{\textbf{ATE [m] on the EuRoC dataset~\cite{euroc}.} In the left table, the results of all methods were obtained by running them on \textbf{stereo} video. In the right table, we report the trajectory error of \textbf{monocular} SLAM. `\xmark’ denotes tracking failure. Results of \cite{orbslam2,orbslam3,svo,droidslam,dso} and \cite{li2023dense} are adopted from \cite{droidslam} and \cite{li2023dense} respectively.}
\label{table:EurocStereo}
\end{table*}

\begin{figure}[!t]
    \centering
    \includegraphics[trim=2cm 15cm 10cm 0cm, clip,width=0.45\textwidth]{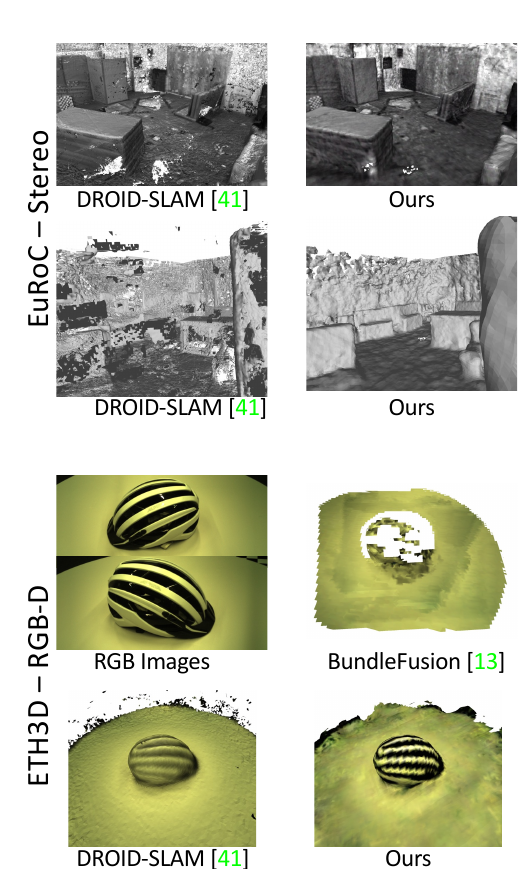}
    \caption{\textbf{Qualitative results.} Top: V1\_02\_Medium (EuRoC), bottom: Helmet (ETH3D). DROID-SLAM~\cite{droidslam} reconstructions obtained with TSDF-Fusion~\cite{tsdffusion}.}
    \label{fig:euroc_eth3d}
\end{figure}

\subsection{Evaluation Metrics} Following the common protocol in the SLAM literature ~\cite{niceslam,orbslam2,droidslam}, we evaluate the estimated trajectory by aligning it to the ground truth and then calculating the camera pose accuracy based on the Absolute Trajectory Error (ATE) RMSE. The reconstruction metrics include \textit{Accuracy} [cm], \textit{Completion} [cm], \textit{Completion Ratio} [$<5$cm \%], and F-score [$<5$cm \%]. For the evaluation, we remove regions not observed by any camera. 
Furthermore, using ground truth trajectory for depth rendering, we evaluate the Depth L1 metric~\cite{niceslam} by computing the absolute error between rendered depths from estimated and ground truth meshes.

\subsection{Comparison with state-of-the-art SLAM}

Here, we evaluate GO-SLAM in synthetic and real-world scenarios and compare it to state-of-the-art SLAM systems in monocular, stereo, and RGB-D settings.  

\textbf{TUM RGB-D (Monocular \& RGB-D).}
In this dataset, we compare with monocular and RGBD SLAM systems. In \cref{table:TUM} (left), we focus on the former methods: traditional SLAM~\cite{orbslam2,orbslam3} with point-based representation fails on camera tracking on most of the challenging sequences. Among all deep-learning-based methods~\cite{deepv2d,deepfactors,droidslam}, GO-SLAM with online LC and full BA achieves the lowest average trajectory error. Furthermore, following~\cite{niceslam}, we also evaluate our method on RGB-D input. \cref{table:TUM} (right) illustrates the clear superiority of our pose estimation compared to recent NeRF-base RGB-D SLAM~\cite{imap,niceslam}, effectively shrinking the gap with traditional SLAM systems.

\begin{figure}[!t]
    \centering
    \includegraphics[width=1.0\columnwidth]{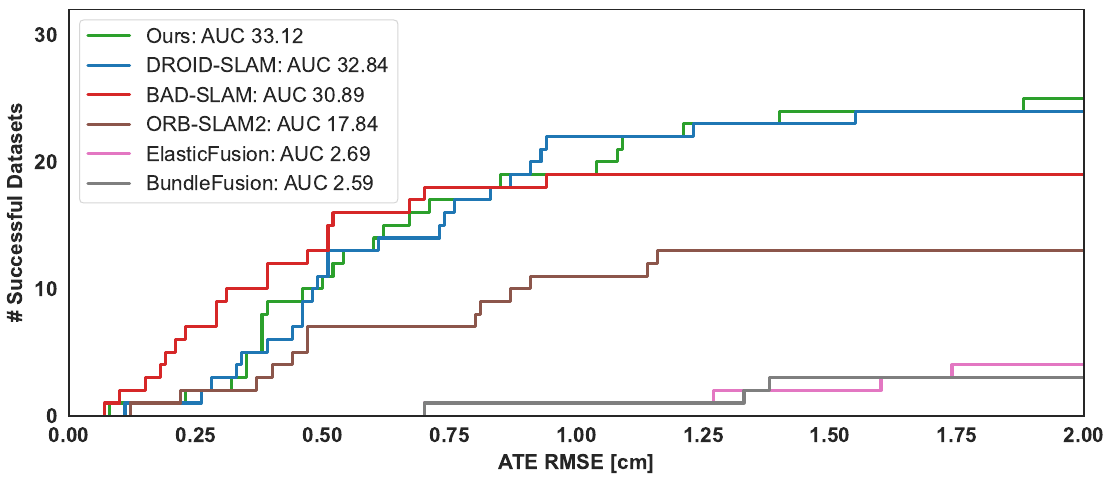}
    \caption{\textbf{RGB-D ETH3D-SLAM test benchmark.} Number of successful trajectories (y-axis) as a function of the ATE RMSE (x-axis). Maximum ATE RMSE: 2cm. Results by published SLAM systems~\cite{droidslam,badslam,orbslam2,elasticfusion,bundlefusion} and ours.
    }
    \label{fig:eth3d}
    % \vspace{-0.2cm}
\end{figure}

\begin{table*}[t!]
  \centering
  \footnotesize
  \setlength{\tabcolsep}{12pt}
  \resizebox{1.0\linewidth}{!}{
    \begin{tabular}{clcccc|cccc|c}
      \toprule
         & Scene ID & \multicolumn{1}{c}{\makecell{\tt{0000}}} &
         \multicolumn{1}{c}{\makecell{\tt{0054}}} &
         \multicolumn{1}{c}{\makecell{\tt{0233}}} &
         \multicolumn{1}{c|}{\makecell{\tt{0465}}} &
         \multicolumn{1}{c}{\makecell{\tt{0059}}} & 
         \multicolumn{1}{c}{\makecell{\tt{0106}}} &   \multicolumn{1}{c}{\makecell{\tt{0169}}} & \multicolumn{1}{c|}{\makecell{\tt{0181}}} & \multirow{2}{*}{Avg.}\\
         \cline{2-10}
         & \# Frames & 5578 & 6629 & 7643 & 6306 & 1807 & 2324 & 2034 & 2349  \\
         \midrule
        \multirow{5}{*}{\rotatebox[origin=c]{90}{RGB-D}} & iMAP$^*$~\cite{imap} & 55.95 & 70.11 & 86.42 & 85.03 & 32.06 & 17.50 &70.51 & 32.10 & 56.21 \\
        & NICE-SLAM~\cite{niceslam} & 8.64 & 20.93 & 9.00 & 22.31 & 12.25 & 8.09 & 10.28 & 12.93 & 13.05 \\
        & DROID-SLAM~\cite{droidslam} (VO) & 8.00 & 29.28 & 6.75 & 11.37 & 11.30 & 9.97 & 8.64 & 7.38 & 11.59 \\
        & DROID-SLAM~\cite{droidslam} & 5.36 & 8.89 & 4.90 & 8.32 & 7.72 & 7.06 & 8.01 & 6.97 & 7.15 \\
        & Ours & \textbf{5.35} & \textbf{8.75} & \textbf{4.78} & \textbf{8.15} & \textbf{7.52} & \textbf{7.03} & \textbf{7.74}  & \textbf{6.84} & \textbf{7.02} \\
        \hline
        \multirow{4}{*}{\rotatebox[origin=c]{90}{Mono.}} & ORB-SLAM3~\cite{orbslam3} & 73.93 & 243.26 & 25.01 & 181.86 & 90.67 & 178.13 & 60.15 & 104.93 & 119.74  \\
        & DROID-SLAM~\cite{droidslam} (VO) & 11.05 & 204.31 & 71.08 & 117.84 & 67.26 & 11.20 & 16.21 & 9.94 & 63.61 \\
        & DROID-SLAM~\cite{droidslam} & \textbf{5.48} & 197.71 & 72.23 & 114.36 & 9.00 & \textbf{6.76} & \textbf{7.86} & \textbf{7.41} & 52.60  \\
        & Ours & 5.94 & \textbf{13.29} & \textbf{5.31} & \textbf{79.51} & \textbf{8.27} & 8.07 & 8.42 & 8.29 & \textbf{17.59}  \\
      \bottomrule
    \end{tabular}
    }%
    \caption{\textbf{ATE[cm] on ScanNet dataset~\cite{scannet}.} For DROID-SLAM~\cite{droidslam}, we also report results for the visual odometry (VO) variant, which does not include the final global bundle adjustment. Results of iMAP$^{*}$ and NICE-SLAM are from~\cite{niceslam}.}
    \label{tab:scannet}
\end{table*}

\begin{table*}[tb!]
  \centering
     \resizebox{1.0\linewidth}{!}{
    \begin{tabular}{lccc|cccccc}
      \toprule
      & \multicolumn{3}{c|}{RGB-D} & \multicolumn{6}{c}{Mono.} \\
      \midrule
       & {iMAP$^*$}~\cite{imap} & NICE-SLAM~\cite{niceslam} & \textbf{Ours} & Orbeez-SLAM~\cite{orbeezslam} & NeRF-SLAM~\cite{nerfslam}$\ddagger$ & DROID-SLAM~\cite{droidslam} & Li \emph{et al.} ~\cite{li2023dense} & NICER-SLAM~\cite{nicerslam} $\ddagger$ &  \textbf{Ours} \\
      \midrule
      \textbf{ATE RMSE[cm]} $\downarrow$ & - & 1.95 & \bf 0.34 & - & - & 0.42 & 0.46 & 1.88 & \bf 0.39 \\
      \midrule
      {\bf Depth L1[cm]} $\downarrow$ & 7.64 & 3.53 & \textbf{3.38} & 11.88 & 4.49 & - & - & - & \textbf{4.39}  \\
      \midrule
      {\bf Acc.[cm]} $\downarrow$ & 6.95 & 2.85 & \textbf{2.50} & - & - & 5.03 & 4.03 & \textbf{3.65} & 3.81 \\
      {\bf Comp.[cm]} $\downarrow$ & 5.33 & \textbf{3.00} & 3.74 & - & - & 8.49 & 4.20 & \textbf{4.16} & 4.79 \\
      {\bf Comp. Ratio[$<$5cm \%]} $\downarrow$ & 66.60 & 89.33 & 88.09 & - & - & 64.72 & \textbf{79.60} & 79.37 & 78.00  \\
      \midrule
      \bf Avg. FPS $\uparrow$ & \textbf{10} & $\ll$ 1 & 8 & $\approx$ 20 & 10 & \textbf{21} & 3 & $\ll$ 1 & 8 \\
       \bottomrule
    \end{tabular}}%
    \caption{\textbf{Reconstruction results and ATE[cm] on the Replica dataset (average over 8 scenes).} The results of iMAP$^*$~\cite{imap} are adopted from NICE-SLAM~\cite{niceslam}, while results for other methods are taken from the respective original papers. $\ddagger$ denotes concurrent works yet unpublished. 
    }  
    \label{tab:replica}
\end{table*}

\textbf{EuRoC (Stereo \& Monocular).}
\cref{table:EurocStereo} (left) shows the camera trajectory error of GO-SLAM for all stereo sequences compared to stereo SLAM methods. 
While existing systems based on neural implicit representation are designed for monocular~\cite{nicerslam,li2023dense,nerfslam,orbeezslam} or RGB-D~\cite{imap,niceslam} input only, our method predicts trajectories comparable to state-of-the-art stereo SLAM~\cite{orbslam2,orbslam3,svo,droidslam}, while also producing globally dense and consistent 3D reconstruction, as shown in \cref{fig:euroc_eth3d}. Compared to the noisy result of DROID-SLAM with several holes and floating points, GO-SLAM produces a more complete, smoother surface and a cleaner reconstruction. In the \cref{table:EurocStereo} (right), we also report the results of monocular SLAM. Specifically, NeRF-based SLAM~\cite{li2023dense} without online BA fails in most cases, while our approach gives accurate predictions in all sequences.

\textbf{ETH3D-SLAM (RGB-D).}
The ETH3D-SLAM dataset provides a public online leaderboard for RGB-D SLAM evaluation. The results in \cref{fig:eth3d} demonstrate the significant advantage of our deep-learning-based method over published RGB-D SLAM systems with point-~\cite{orbslam2}, surfel-~\cite{badslam,elasticfusion} or voxel-based~\cite{bundlefusion} representations. Similarly to our approach, BundleFusion~\cite{bundlefusion} targets globally consistent reconstruction. However, its pose estimation is highly susceptible to errors, resulting in inferior reconstruction as shown in \cref{fig:euroc_eth3d}, whereas our method yields smoother reconstruction with photometrically convincing rendering.

\begin{table}[t]
    \centering
    \resizebox{0.75\columnwidth}{!}{
    \begin{tabular}{ccc|c}
    \toprule
    \multicolumn{3}{c|}{Sampling Strategy} & Avg. \\
    \cline{1-3}
    Latest & Stratified & Top-Ranked & F-score $\uparrow$ \\
    \midrule
    \cmark & \xmark & \xmark & 49.55 \\
    \cmark & \cmark & \xmark & 83.13 \\
    \cmark & \cmark & \cmark & \bf 85.56 \\
    \bottomrule
    \end{tabular}
    }
    \caption{\textbf{Impact of keyframe selection.} We report average F-score achieved by different sampling strategies.}
    \label{tab:ablate_kf_sel}
\end{table}

\textbf{ScanNet (Monocular \& RGB-D).}
For an exhaustive evaluation, we test SLAM methods on both short (sequences with less than 5000 frames) and long, complex sequences. \cref{tab:scannet} shows that, although DROID-SLAM~\cite{droidslam} performs well on short sequences and RGB-D inputs, its accuracy drops dramatically when processing longer sequences in the monocular setting, even performing a final global bundle adjustment. In contrast, GO-SLAM consistently yields the best results, thanks to online loop closing and full BA. Furthermore, as anticipated in \cref{fig:teaser}, due to the absence of global optimization to eliminate the accumulated trajectory error, NeRF-based SLAM, \eg, iMAP, and NICE-SLAM~\cite{niceslam} provide significantly poorer results for pose estimation and 3D reconstruction.

\textbf{Replica (Monocular \& RGB-D).} 
Finally, \cref{tab:replica} shows pose and reconstruction performance achieved by GO-SLAM on the Replica dataset~\cite{replica}. Our system achieves comparable accuracy with respect to existing RGB-D methods~\cite{niceslam} and concurrent monocular~\cite{nicerslam,li2023dense} ones. However, none of the latter runs in real-time, whereas GO-SLAM achieves a frame rate of 8 FPS with maximum GPU memory consuming 18 GB. Although iMAP~\cite{imap} alone can achieve a similar speed as GO-SLAM (10 FPS), it yields much worse results. In particular, close to ours, NeRF-SLAM~\cite{nerfslam} is a concurrent, monocular SLAM system based on DROID-SLAM~\cite{droidslam}, running at nearly 10 FPS. However, their system lacks BA, which GO-SLAM provides. Moreover, NeRF-SLAM and Orbeez-SLAM~\cite{orbeezslam} lack dense 3D reconstruction results, limiting the comparison. We can only compare on depth rendering quality, for which GO-SLAM achieves better results. 

\subsection{Ablation Study}
We conclude by studying the impact of the novel designs in the proposed GO-SLAM with RGB-D input.

\textbf{Keyframe Selection.}
In \cref{tab:ablate_kf_sel}, we investigate the impact of different keyframe selection strategies by computing the mean F-score in eight sequences of the Replica dataset~\cite{replica}. The experimental results prove that updating the 3D model with the latest keyframes or stratified sampling is not sufficient to obtain the best results. 
Monitoring the pose and depth of each frame continuously and updating the 3D model according to the largest change is crucial for globally consistent reconstruction.

\begin{table}[t]
    \centering
    \resizebox{0.7\columnwidth}{!}{
    \begin{tabular}{cccc|c}
    \toprule
    $\mathcal{L}_{c}$ & $\mathcal{L}_{dep}$ & $\mathcal{L}_{sdf}$ & $\mathcal{L}_{eik}$ & Avg. F-score $\uparrow$ \\
    \midrule
    \cmark & \xmark & \xmark & \xmark & 34.64 \\
    \cmark & \cmark & \xmark & \xmark & 83.50 \\
    \cmark & \xmark & \cmark & \xmark & 84.00 \\
    \cmark & \cmark & \cmark & \xmark & 84.83 \\
    \cmark & \cmark & \cmark & \cmark & \bf 85.56 \\
    \bottomrule
    \end{tabular}
    }
    \caption{\textbf{Impact of single losses.} We report average F-score achieved with different losses configurations.}
    \label{tab:ablate_loss}
\end{table}

\begin{table}[t]
    \centering
    \resizebox{0.8\columnwidth}{!}{
    \begin{tabular}{c|c|cc}
    \toprule
    Skipping & \multirow{2}{*}{Speedup} & Avg. & Avg. ATE \\
    Frames & & F-score $\uparrow$ & RMSE [cm] $\downarrow$ \\
    \midrule
    None & $1\times$ & \bf 85.56 & \bf 7.02 \\
    1/2 & $2\times$ & 84.67 & 7.08 \\
    3/4 & $4\times$ & 84.80 & 7.16 \\
    7/8 & $8\times$ & 84.41 & 7.28 \\
    \bottomrule
    \end{tabular}
    }
    \caption{\textbf{Impact of frames skipping.} We report average F-score and ATE when running in real-time.}
    \label{tab:ablate_skipping}
\end{table}

\textbf{Losses.}
\cref{tab:ablate_loss} evaluates on Replica the effectiveness of each loss term.
Results show that the RGB loss alone cannot lead to satisfying results. Geometric supervision, such as depth or SDF loss, provides similar accuracy while integrating all terms produces the best results.

\textbf{Skipping frames to run in real-time.} \cref{tab:ablate_skipping} shows our GO-SLAM accuracy when skipping RGB-D frames to run at $2\times, 4\times, 8\times$ speed. Remarkably, both mapping accuracy (avg. F-score on Replica~\cite{replica}) and tracking performance (avg. ATE RMSE on ScanNet~\cite{scannet}) only experience minimal degradation, proving that GO-SLAM is 1) robust to large view changes and 2) ready for real-time usage.

\textbf{Loop Closing and Full BA.}
Finally, we study the impact of our efficient loop closing and online full BA on eight scenes of ScanNet~\cite{scannet}. 
Our baseline DROID-SLAM (VO)~\cite{droidslam} (without LC and Full BA) achieves high speed but suffers from a large trajectory error, as shown in \cref{tab:ablate_lc_fba}. Our efficient LC significantly reduces drift with negligible speed reduction. Introducing Full BA slows down GO-SLAM, but improves global pose estimation significantly.
Finally, our full system integrating LC and full BA achieves the best results in pose estimation and 3D reconstruction (see \cref{fig:ba}), while enabling real-time performance.

\subsection{CPU/GPU Requirements.}

We conclude by measuring hardware requirements on the Replica dataset~\cite{replica}. In \cref{tab:memory} we present several metrics including CPU requirements, maximum GPU memory consumption, average frames per second, and final accuracy, showcasing the performance of various SLAM algorithms during the reconstruction of an entire scene.
Specifically, our SLAM system stands out by achieving an optimal balance between CPU/GPU requirements and SLAM performance, combining high accuracy with speed.

\begin{table}[t]
    \centering
    \resizebox{0.9\columnwidth}{!}{
    \begin{tabular}{ccc}
    \toprule
    & Avg. ATE RMSE [cm] $\downarrow$ & avg. FPS $\uparrow$ \\
    \midrule
    w/o LC \&  w/o Full BA & 11.59 & \textbf{30} \\
    w/ LC & 8.83 & 20 \\
    w/ Full BA & 7.11 & 12 \\
    w/ LC \& w/ Full BA & \textbf{7.02} & 10 \\
    \bottomrule
    \end{tabular}
    }
    \caption{\textbf{Impact of loop closing and full BA.} We report average ATE and FPS with/without our main contributions.}
    \label{tab:ablate_lc_fba}
\end{table}

\begin{table}[t]
    \centering
    \resizebox{1.0\columnwidth}{!}{
    \begin{tabular}{c|cccc}
    \toprule
    \multirow{2}{*}{Method}& CPU Processing & GPU Consuming & Avg. & Comp.\\
    & Frequency [GHz] & Memory [G]$\downarrow$ & FPS$\uparrow$ & Ratio[$<$5cm \%] $\uparrow$ \\
    \midrule
    iMAP$^{*}$~\cite{imap} & 3.80 & \bf 10.13 & 10 & 66.60 \\
    NICE-SLAM~\cite{niceslam} & 3.80 & 11.72 & $\ll$ 1 & \bf 89.33 \\
    DROID-SLAM~\cite{droidslam} & 3.50 & 14.34 & \bf 21 & 70.52 \\
    Ours & 3.50 & 15.63 & 8 & 88.09 \\
    \bottomrule
    \end{tabular}
    }  
    \caption{\textbf{Hardware requirements and performance.} All results are obtained by running on NVIDIA RTX 3090 GPU and Replica dataset~\cite{replica} with RGB-D input.}
    \label{tab:memory}
\end{table}
\section{Conclusions}

We introduced a novel, real-time deep-learning-based SLAM algorithm that achieves globally consistent reconstruction with monocular, stereo, or RGB-D input. Our approach explicitly detects loop closures and performs online full BA to minimize trajectory error. Based on NeRF, our 3D reconstruction provides an efficient, compact, and multi-resolution representation. Moreover, our approach continuously updates the dense 3D reconstruction to adapt to the newly-optimized global geometry at high frequency. Our experiments demonstrate the algorithm's robustness in reliably tracking and densely mapping even in large-scale scenes, especially on long monocular trajectories with no depth information, achieving state-of-the-art performance on various datasets.

\textbf{Acknowledgment.} We sincerely thank the scholarship supported by China Scholarship Council (CSC).

{\small
\bibliographystyle{ieee}
\bibliography{egbib}

\begin{thebibliography}{10}\itemsep=-1pt

\bibitem{neuralsurface}
Dejan Azinovi{\'c}, Ricardo Martin-Brualla, Dan~B Goldman, Matthias
  Nie{\ss}ner, and Justus Thies.
\newblock Neural rgb-d surface reconstruction.
\newblock In {\em CVPR}, pages 6290--6301, 2022.

\bibitem{mipnerf}
Jonathan~T Barron, Ben Mildenhall, Matthew Tancik, Peter Hedman, Ricardo
  Martin-Brualla, and Pratul~P Srinivasan.
\newblock Mip-nerf: A multiscale representation for anti-aliasing neural
  radiance fields.
\newblock In {\em ICCV}, pages 5855--5864, 2021.

\bibitem{mipnerf360}
Jonathan~T Barron, Ben Mildenhall, Dor Verbin, Pratul~P Srinivasan, and Peter
  Hedman.
\newblock Mip-nerf 360: Unbounded anti-aliased neural radiance fields.
\newblock In {\em CVPR}, pages 5470--5479, 2022.

\bibitem{codeslam}
Michael Bloesch, Jan Czarnowski, Ronald Clark, Stefan Leutenegger, and Andrew~J
  Davison.
\newblock Codeslam—learning a compact, optimisable representation for dense
  visual slam.
\newblock In {\em CVPR}, pages 2560--2568, 2018.

\bibitem{euroc}
Michael Burri, Janosch Nikolic, Pascal Gohl, Thomas Schneider, Joern Rehder,
  Sammy Omari, Markus~W Achtelik, and Roland Siegwart.
\newblock The euroc micro aerial vehicle datasets.
\newblock {\em IJRR}, 35(10):1157--1163, 2016.

\bibitem{orbslam3}
Carlos Campos, Richard Elvira, Juan J~G{\'o}mez Rodr{\'\i}guez, Jos{\'e}~MM
  Montiel, and Juan~D Tard{\'o}s.
\newblock Orb-slam3: An accurate open-source library for visual,
  visual-inertial and multi-map slam.
\newblock {\em arXiv preprint arXiv:2007.11898}, 2020.

\bibitem{rcmvsnet}
Di Chang, Alja{\v{z}} Bo{\v{z}}i{\v{c}}, Tong Zhang, Qingsong Yan, Yingcong
  Chen, Sabine S{\"u}sstrunk, and Matthias Nie{\ss}ner.
\newblock Rc-mvsnet: unsupervised multi-view stereo with neural rendering.
\newblock In {\em ECCV}, pages 665--680. Springer, 2022.

\bibitem{mvsnerf}
Anpei Chen, Zexiang Xu, Fuqiang Zhao, Xiaoshuai Zhang, Fanbo Xiang, Jingyi Yu,
  and Hao Su.
\newblock Mvsnerf: Fast generalizable radiance field reconstruction from
  multi-view stereo.
\newblock In {\em ICCV}, pages 14124--14133, 2021.

\bibitem{orbeezslam}
Chi-Ming Chung, Yang-Che Tseng, Ya-Ching Hsu, Xiang-Qian Shi, Yun-Hung Hua,
  Jia-Fong Yeh, Wen-Chin Chen, Yi-Ting Chen, and Winston~H Hsu.
\newblock Orbeez-slam: A real-time monocular visual slam with orb features and
  nerf-realized mapping.
\newblock {\em arXiv preprint arXiv:2209.13274}, 2022.

\bibitem{tsdffusion}
Brian Curless and Marc Levoy.
\newblock A volumetric method for building complex models from range images.
\newblock In {\em SIGGRAPH}, pages 303--312, 1996.

\bibitem{deepfactors}
Jan Czarnowski, Tristan Laidlow, Ronald Clark, and Andrew~J Davison.
\newblock Deepfactors: Real-time probabilistic dense monocular slam.
\newblock {\em IEEE RAL}, 5(2):721--728, 2020.

\bibitem{scannet}
Angela Dai, Angel~X Chang, Manolis Savva, Maciej Halber, Thomas Funkhouser, and
  Matthias Nie{\ss}ner.
\newblock Scannet: Richly-annotated 3d reconstructions of indoor scenes.
\newblock In {\em CVPR}, pages 5828--5839, 2017.

\bibitem{bundlefusion}
Angela Dai, Matthias Nie{\ss}ner, Michael Zollh{\"o}fer, Shahram Izadi, and
  Christian Theobalt.
\newblock Bundlefusion: Real-time globally consistent 3d reconstruction using
  on-the-fly surface reintegration.
\newblock {\em ACM TOG}, 36(4):1, 2017.

\bibitem{dso}
Jakob Engel, Vladlen Koltun, and Daniel Cremers.
\newblock Direct sparse odometry.
\newblock {\em IEEE TPAMI}, 40(3):611--625, 2017.

\bibitem{svo}
Christian Forster, Zichao Zhang, Michael Gassner, Manuel Werlberger, and Davide
  Scaramuzza.
\newblock Svo: Semidirect visual odometry for monocular and multicamera
  systems.
\newblock {\em IEEE Transactions on Robotics}, 33(2):249--265, 2016.

\bibitem{plenoxels}
Sara Fridovich-Keil, Alex Yu, Matthew Tancik, Qinhong Chen, Benjamin Recht, and
  Angjoo Kanazawa.
\newblock Plenoxels: Radiance fields without neural networks.
\newblock In {\em CVPR}, pages 5501--5510, 2022.

\bibitem{gallup20103d}
David Gallup, Marc Pollefeys, and Jan-Michael Frahm.
\newblock 3d reconstruction using an n-layer heightmap.
\newblock In {\em PR}, pages 1--10. Springer, 2010.

\bibitem{eikonal}
Amos Gropp, Lior Yariv, Niv Haim, Matan Atzmon, and Yaron Lipman.
\newblock Implicit geometric regularization for learning shapes.
\newblock {\em arXiv preprint arXiv:2002.10099}, 2020.

\bibitem{di-fusion}
Jiahui Huang, Shi-Sheng Huang, Haoxuan Song, and Shi-Min Hu.
\newblock Di-fusion: Online implicit 3d reconstruction with deep priors.
\newblock In {\em CVPR}, pages 8932--8941, 2021.

\bibitem{li2023dense}
Heng Li, Xiaodong Gu, Weihao Yuan, Luwei Yang, Zilong Dong, and Ping Tan.
\newblock Dense rgb slam with neural implicit maps.
\newblock {\em ICLR}, 2023.

\bibitem{bnvfusion}
Kejie Li, Yansong Tang, Victor~Adrian Prisacariu, and Philip~HS Torr.
\newblock Bnv-fusion: dense 3d reconstruction using bi-level neural volume
  fusion.
\newblock In {\em CVPR}, pages 6166--6175, 2022.

\bibitem{barf}
Chen-Hsuan Lin, Wei-Chiu Ma, Antonio Torralba, and Simon Lucey.
\newblock Barf: Bundle-adjusting neural radiance fields.
\newblock In {\em ICCV}, pages 5741--5751, 2021.

\bibitem{nerf}
Ben Mildenhall, Pratul~P Srinivasan, Matthew Tancik, Jonathan~T Barron, Ravi
  Ramamoorthi, and Ren Ng.
\newblock Nerf: Representing scenes as neural radiance fields for view
  synthesis.
\newblock In {\em ECCV}, pages 405--421, 2020.

\bibitem{instantngp}
Thomas M{\"u}ller, Alex Evans, Christoph Schied, and Alexander Keller.
\newblock Instant neural graphics primitives with a multiresolution hash
  encoding.
\newblock {\em ACM TOG}, 41(4):1--15, 2022.

\bibitem{orbslam}
Raul Mur-Artal, Jose Maria~Martinez Montiel, and Juan~D Tardos.
\newblock Orb-slam: a versatile and accurate monocular slam system.
\newblock {\em IEEE Transactions on Robotics}, 31(5):1147--1163, 2015.

\bibitem{orbslam2}
Raul Mur-Artal and Juan~D Tard{\'o}s.
\newblock Orb-slam2: An open-source slam system for monocular, stereo, and
  rgb-d cameras.
\newblock {\em IEEE Transactions on Robotics}, 33(5):1255--1262, 2017.

\bibitem{nyuv2}
Pushmeet~Kohli Nathan~Silberman, Derek~Hoiem and Rob Fergus.
\newblock Indoor segmentation and support inference from rgbd images.
\newblock In {\em ECCV}, 2012.

\bibitem{kinectfusion}
Richard~A Newcombe, Shahram Izadi, Otmar Hilliges, David Molyneaux, David Kim,
  Andrew~J Davison, Pushmeet Kohi, Jamie Shotton, Steve Hodges, and Andrew
  Fitzgibbon.
\newblock Kinectfusion: Real-time dense surface mapping and tracking.
\newblock In {\em IEEE ISMAR}, pages 127--136. IEEE, 2011.

\bibitem{voxelhashing}
Matthias Nie{\ss}ner, Michael Zollh{\"o}fer, Shahram Izadi, and Marc
  Stamminger.
\newblock Real-time 3d reconstruction at scale using voxel hashing.
\newblock {\em ACM TOG}, 32(6):1--11, 2013.

\bibitem{isdf}
Joseph Ortiz, Alexander Clegg, Jing Dong, Edgar Sucar, David Novotny, Michael
  Zollhoefer, and Mustafa Mukadam.
\newblock isdf: Real-time neural signed distance fields for robot perception.
\newblock {\em arXiv preprint arXiv:2204.02296}, 2022.

\bibitem{nerfslam}
Antoni Rosinol, John~J Leonard, and Luca Carlone.
\newblock Nerf-slam: Real-time dense monocular slam with neural radiance
  fields.
\newblock {\em arXiv preprint arXiv:2210.13641}, 2022.

\bibitem{badslam}
Thomas Schops, Torsten Sattler, and Marc Pollefeys.
\newblock Bad slam: Bundle adjusted direct rgb-d slam.
\newblock In {\em CVPR}, pages 134--144, 2019.

\bibitem{replica}
Julian Straub, Thomas Whelan, Lingni Ma, Yufan Chen, Erik Wijmans, Simon Green,
  Jakob~J Engel, Raul Mur-Artal, Carl Ren, Shobhit Verma, et~al.
\newblock The replica dataset: A digital replica of indoor spaces.
\newblock {\em arXiv preprint arXiv:1906.05797}, 2019.

\bibitem{tum}
J{\"u}rgen Sturm, Nikolas Engelhard, Felix Endres, Wolfram Burgard, and Daniel
  Cremers.
\newblock A benchmark for the evaluation of rgb-d slam systems.
\newblock In {\em 2012 IEEE/RSJ international conference on intelligent robots
  and systems}, pages 573--580. IEEE, 2012.

\bibitem{imap}
Edgar Sucar, Shikun Liu, Joseph Ortiz, and Andrew~J Davison.
\newblock imap: Implicit mapping and positioning in real-time.
\newblock In {\em ICCV}, pages 6229--6238, 2021.

\bibitem{dvgo}
Cheng Sun, Min Sun, and Hwann-Tzong Chen.
\newblock Direct voxel grid optimization: Super-fast convergence for radiance
  fields reconstruction.
\newblock In {\em CVPR}, pages 5459--5469, 2022.

\bibitem{neural3dwild}
Jiaming Sun, Xi Chen, Qianqian Wang, Zhengqi Li, Hadar Averbuch-Elor, Xiaowei
  Zhou, and Noah Snavely.
\newblock Neural 3d reconstruction in the wild.
\newblock In {\em SIGGRAPH}, pages 1--9, 2022.

\bibitem{cnnslam}
Keisuke Tateno, Federico Tombari, Iro Laina, and Nassir Navab.
\newblock Cnn-slam: Real-time dense monocular slam with learned depth
  prediction.
\newblock In {\em CVPR}, pages 6243--6252, 2017.

\bibitem{deepv2d}
Zachary Teed and Jia Deng.
\newblock Deepv2d: Video to depth with differentiable structure from motion.
\newblock {\em arXiv preprint arXiv:1812.04605}, 2018.

\bibitem{raft}
Zachary Teed and Jia Deng.
\newblock Raft: Recurrent all-pairs field transforms for optical flow.
\newblock In {\em ECCV}, pages 402--419. Springer, 2020.

\bibitem{droidslam}
Zachary Teed and Jia Deng.
\newblock Droid-slam: Deep visual slam for monocular, stereo, and rgb-d
  cameras.
\newblock {\em NeurIPS}, 34:16558--16569, 2021.

\bibitem{gosurf}
Jingwen Wang, Tymoteusz Bleja, and Lourdes Agapito.
\newblock Go-surf: Neural feature grid optimization for fast, high-fidelity
  rgb-d surface reconstruction.
\newblock {\em arXiv preprint arXiv:2206.14735}, 2022.

\bibitem{neus}
Peng Wang, Lingjie Liu, Yuan Liu, Christian Theobalt, Taku Komura, and Wenping
  Wang.
\newblock Neus: Learning neural implicit surfaces by volume rendering for
  multi-view reconstruction.
\newblock {\em arXiv preprint arXiv:2106.10689}, 2021.

\bibitem{ibrnet}
Qianqian Wang, Zhicheng Wang, Kyle Genova, Pratul~P Srinivasan, Howard Zhou,
  Jonathan~T Barron, Ricardo Martin-Brualla, Noah Snavely, and Thomas
  Funkhouser.
\newblock Ibrnet: Learning multi-view image-based rendering.
\newblock In {\em CVPR}, pages 4690--4699, 2021.

\bibitem{elasticfusion}
Thomas Whelan, Stefan Leutenegger, R Salas-Moreno, Ben Glocker, and Andrew
  Davison.
\newblock Elasticfusion: Dense slam without a pose graph.
\newblock Robotics: Science and Systems, 2015.

\bibitem{hrbf}
Yabin Xu, Liangliang Nan, Laishui Zhou, Jun Wang, and Charlie~CL Wang.
\newblock Hrbf-fusion: Accurate 3d reconstruction from rgb-d data using
  on-the-fly implicits.
\newblock {\em ACM TOG}, 41(3):1--19, 2022.

\bibitem{inerf}
Lin Yen-Chen, Pete Florence, Jonathan~T Barron, Alberto Rodriguez, Phillip
  Isola, and Tsung-Yi Lin.
\newblock inerf: Inverting neural radiance fields for pose estimation.
\newblock In {\em IROS}, pages 1323--1330. IEEE, 2021.

\bibitem{Yu2022MonoSDF}
Zehao Yu, Songyou Peng, Michael Niemeyer, Torsten Sattler, and Andreas Geiger.
\newblock Monosdf: Exploring monocular geometric cues for neural implicit
  surface reconstruction.
\newblock {\em NeurIPS}, 2022.

\bibitem{nerf++}
Kai Zhang, Gernot Riegler, Noah Snavely, and Vladlen Koltun.
\newblock Nerf++: Analyzing and improving neural radiance fields.
\newblock {\em arXiv preprint arXiv:2010.07492}, 2020.

\bibitem{nerfusion}
Xiaoshuai Zhang, Sai Bi, Kalyan Sunkavalli, Hao Su, and Zexiang Xu.
\newblock Nerfusion: Fusing radiance fields for large-scale scene
  reconstruction.
\newblock In {\em CVPR}, pages 5449--5458, 2022.

\bibitem{deeptam}
Huizhong Zhou, Benjamin Ummenhofer, and Thomas Brox.
\newblock Deeptam: Deep tracking and mapping.
\newblock In {\em ECCV}, pages 822--838, 2018.

\bibitem{nicerslam}
Zihan Zhu, Songyou Peng, Viktor Larsson, Zhaopeng Cui, Martin~R Oswald, Andreas
  Geiger, and Marc Pollefeys.
\newblock Nicer-slam: Neural implicit scene encoding for rgb slam.
\newblock {\em arXiv preprint arXiv:2302.03594}, 2023.

\bibitem{niceslam}
Zihan Zhu, Songyou Peng, Viktor Larsson, Weiwei Xu, Hujun Bao, Zhaopeng Cui,
  Martin~R Oswald, and Marc Pollefeys.
\newblock Nice-slam: Neural implicit scalable encoding for slam.
\newblock In {\em CVPR}, pages 12786--12796, 2022.

\bibitem{mononeuralfusion}
Zi-Xin Zou, Shi-Sheng Huang, Yan-Pei Cao, Tai-Jiang Mu, Ying Shan, and Hongbo
  Fu.
\newblock Mononeuralfusion: Online monocular neural 3d reconstruction with
  geometric priors.
\newblock {\em arXiv preprint arXiv:2209.15153}, 2022.

\end{thebibliography}
}

\newpage\phantom{Supplementary}
\multido{\i=1+1}{4}{


\section*{Supplementary material}

# GO-SLAM: Global Optimization for Consistent 3D Instant Reconstruction – Supplementary Material

Youmin Zhang     Fabio Tosi     Stefano Mattoccia     Matteo Poggi

Department of Computer Science and Engineering (DISI)
University of Bologna, Italy
{youmin.zhang2, fabio.tosi5, stefano.mattoccia, m.poggi}@unibo.it
https://youmi-zym.github.io/projects/GO-SLAM/

## A. SLAM Configuration

### A.1. Rendering Network Details.

**Architecture.** Our rendering networks consist of one SDF network and one color network as shown in Fig. A. For multi-resolution hash encodings, we set the feature dimensions and maximum entries per level to 2 and $2^{19}$ respectively. Totally, there are 16 levels of different resolutions ranging from 16 to 4096. The size of hidden units in the SDF network and color network are 32 and 64 respectively.

**Training.** We use AdamW [4] as optimizer. The learning rate is set to $10^{-2}$ for learnable hash encodings and $10^{-3}$ for parameters of the SDF network and color network.

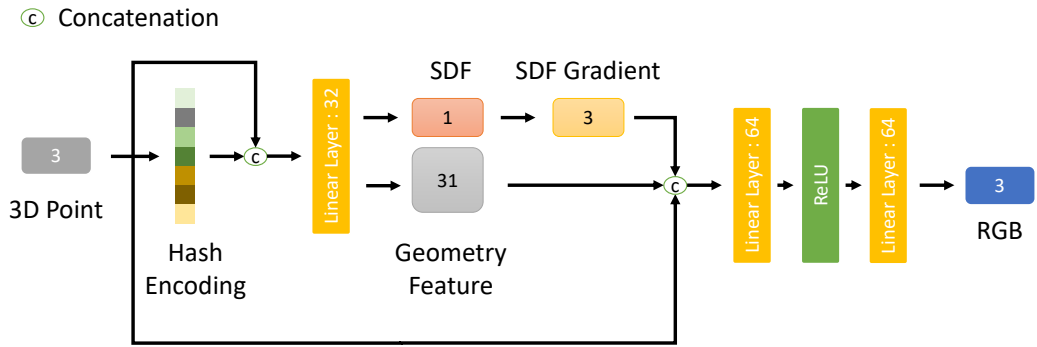

Figure A: **Details of rendering networks.**

## B. More Qualitative Results

We provide detailed results on ScanNet [3], EuRoC [1] and Replica [5] datasets on Figs. B to D respectively.

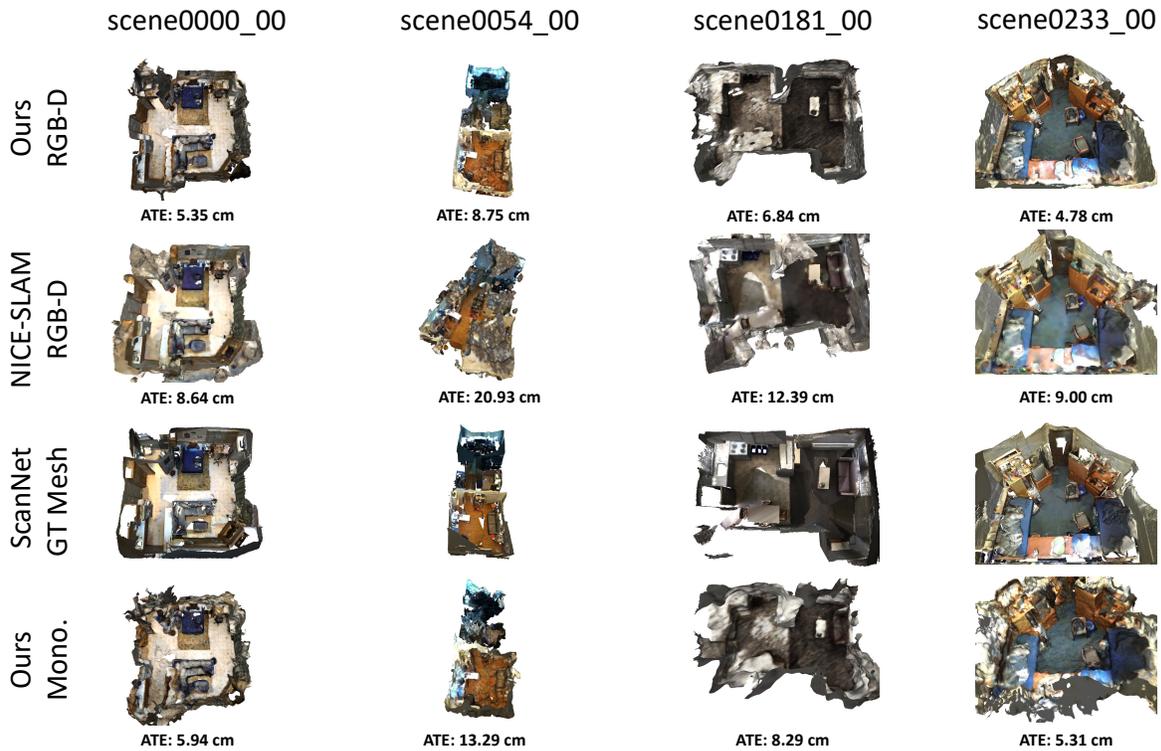

Figure B: **Qualitative results on ScanNet dataset [5].** Our GO-SLAM predicts globally consistent 3D reconstruction with monocular or RGB-D input at real-time speed. While NICE-SLAM [7] suffers from accumulated errors in trajectory and distortion in 3D models.

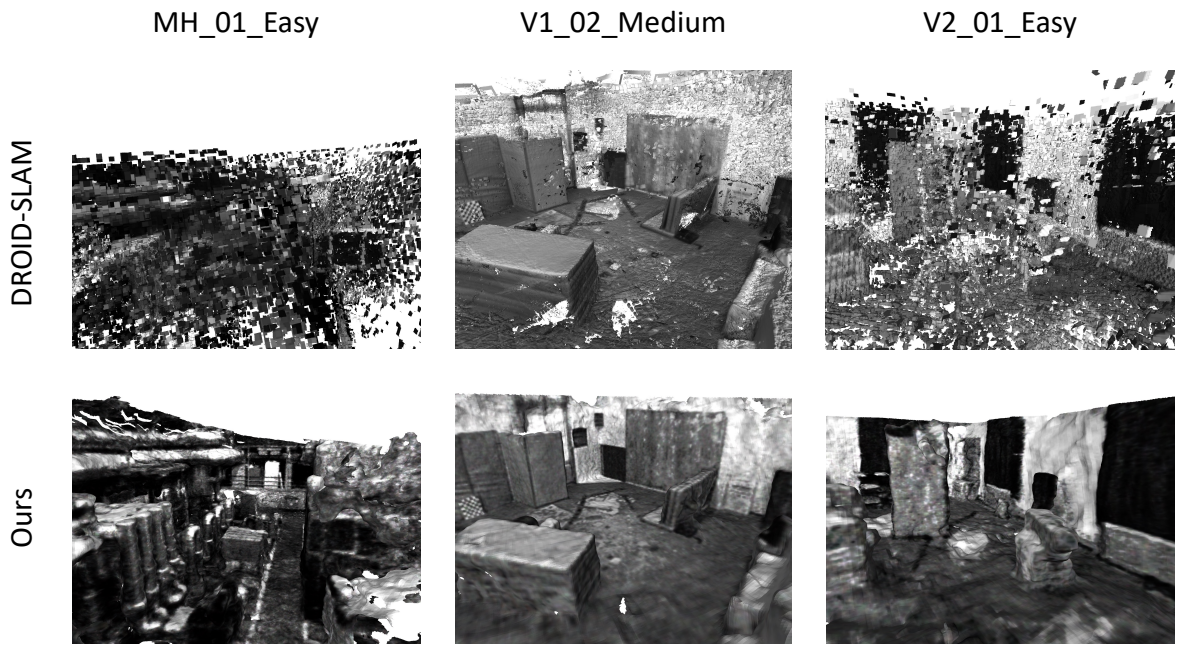

Figure C: **Qualitative results on EuRoC dataset [1].** DROID-SLAM [6] reconstructions obtained with TSDF-Fusion [2]. Compared to DROID-SLAM, our reconstructions are much cleaner and more complete.

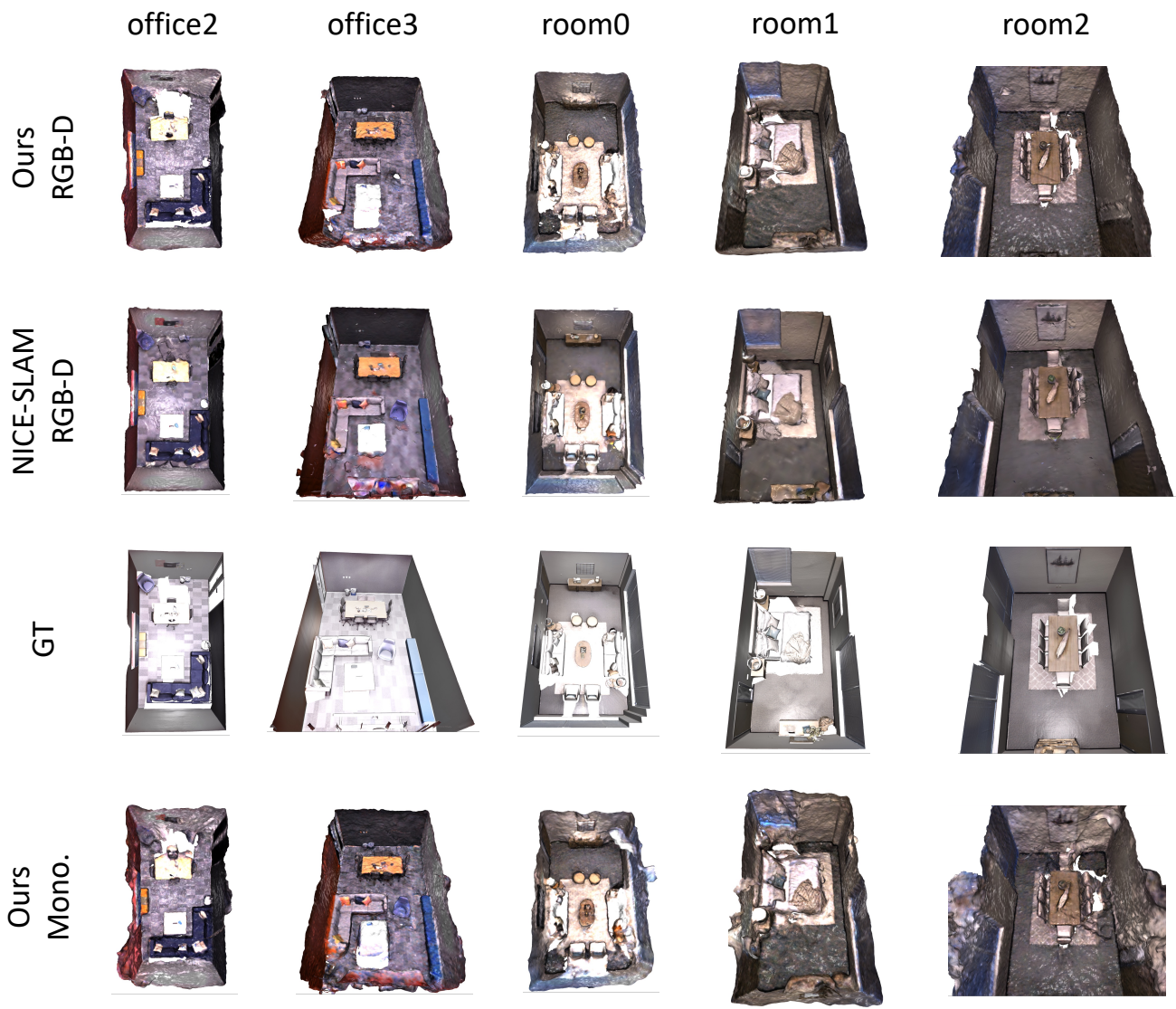

Figure D: **Qualitative results on Replica dataset [5].** Our GO-SLAM predicts dense 3D reconstruction with monocular or RGB-D input at real-time speed. While NICE-SLAM [7], which can only work with depth input, takes hours ($\ll$ 1 FPS) to get the reconstruction.


}

\end{document}